\documentclass[lettersize,journal]{IEEEtran}
\usepackage{amsmath,amsfonts}
\usepackage{algorithmic}
\usepackage{algorithm}
\usepackage{array}
\usepackage{textcomp}
\usepackage{stfloats}
\usepackage{url}
\usepackage{verbatim}
\usepackage{graphicx}
\usepackage{subfigure}
\usepackage[compress]{cite} 
\usepackage{float}
\usepackage{booktabs}
\usepackage{multirow}
\usepackage{makecell}
\usepackage{tikz,xcolor}
\usepackage[colorlinks,linkcolor=blue]{hyperref}
\hypersetup{hidelinks,
	colorlinks=true,
	allcolors=black,
	pdfstartview=Fit,
	breaklinks=true}
\definecolor{lime}{HTML}{A6CE39}

\begin{document}
\title{Multimodal Large Language Models for End-to-End Affective Computing: Benchmarking and Boosting with Generative Knowledge Prompting}

\author{
	\IEEEauthorblockN{
		Miaosen Luo\IEEEauthorrefmark{1}, 
		Jiesen Long\IEEEauthorrefmark{1},
		Zequn Li\IEEEauthorrefmark{1}, 
		Yunying Yang\IEEEauthorrefmark{3}, 
		Yuncheng Jiang\IEEEauthorrefmark{1},
        Sijie Mai\thanks{† Corresponding author}\IEEEauthorrefmark{1}\IEEEauthorrefmark{2}
        }
        
	\IEEEauthorblockA{\IEEEauthorrefmark{1}School of Computer Science, South China Normal University \\ Guangzhou, Guangdong, China\\}
    \IEEEauthorblockA{\IEEEauthorrefmark{3}School of Information Technology in Education, South China Normal University \\ Guangzhou, Guangdong, China\\}
    \IEEEauthorblockA{\IEEEauthorrefmark{2}sijiemai@m.scnu.edu.cn}
}

\maketitle
\begin{abstract}
    Multimodal Affective Computing (MAC) aims to recognize and interpret human emotions by integrating information from diverse modalities such as text, video, and audio. Recent advancements in Multimodal Large Language Models (MLLMs) have significantly reshaped the landscape of MAC by offering a unified framework for processing and aligning cross-modal information. However, practical challenges remain, including performance variability across complex MAC tasks and insufficient understanding of how architectural designs and data characteristics impact affective analysis. To address these gaps, we conduct a systematic benchmark evaluation of state-of-the-art open-source MLLMs capable of concurrently processing audio, visual, and textual modalities across multiple established MAC datasets. Our evaluation not only compares the performance of these MLLMs but also provides actionable insights into model optimization by analyzing the influence of model architectures and dataset properties. Furthermore, we propose a novel hybrid strategy that combines generative knowledge prompting with supervised fine-tuning to enhance MLLMs' affective computing capabilities. Experimental results demonstrate that this integrated approach significantly improves performance across various MAC tasks, offering a promising avenue for future research and development in this field. Our code is released on https://github.com/LuoMSen/MLLM-MAC.
\end{abstract}

\section{Introduction}
Multimodal Affective Computing (MAC) aims to recognize, perceive, infer, and interpret human emotions through the integration of information from multiple modalities, including text, video, and audio \cite{poria2017review}. Human emotional expressions are inherently complex and multimodal in nature \cite{wang2022systematic}, a characteristic that makes unimodal approaches particularly vulnerable to ambiguity, noise interference, and information loss \cite{sanku2024effective}. By fusing data across diverse modalities, multimodal affective computing effectively mitigates this limitation, enabling a more comprehensive and robust understanding of affective states—an approach that aligns with the holistic manner in which humans perceive emotions \cite{li2024surveying}. In prior studies \cite{mai2025learning}, the majority of works employ non-end-to-end approaches, where pre-processed features are used for model construction. While this method has achieved partial success, it considerably restricts the model's ability to deeply explore and automatically learn emotional information. Thus, exploring and developing end-to-end multimodal affective computing approaches has become a key trend in current research.

\begin{figure}[h]
    \centering
    \includegraphics[width=0.48\textwidth]{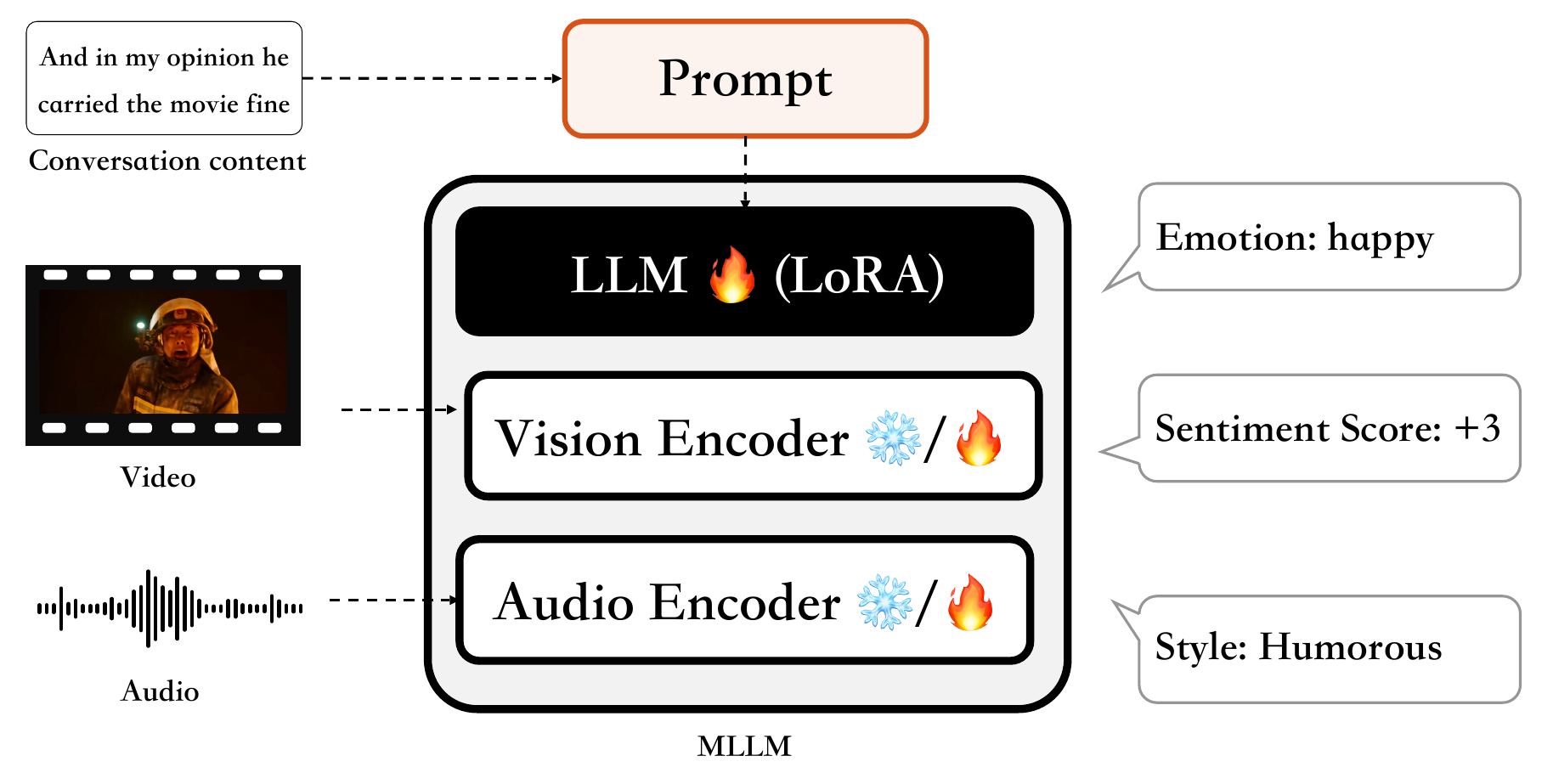} 
    \caption{Schematic diagram of the end-to-end affective computing process based on the multimodal large language model (MLLM). In this process, the original video and audio data are directly fed into the MLLM, while the conversational text is embedded into the prompt, thereby outputting the corresponding affective computing results.}
    \label{fig:MLLM end-to-end}
\end{figure}

The recent revolutionary advancements in Large Language Models (LLMs) have dramatically reshaped the landscape of natural language processing and beyond \cite{gao2025llm}. Their extraordinary ability to comprehend, reason, and generate human-like text originates from extensive pre-training on massive corpora. Importantly, this progress has rapidly expanded to Multimodal Large Language Models (MLLMs), such as GPT-4V \cite{yang2023dawn}, LLaVA \cite{liu2023visual}, Gemini \cite{team2023gemini}, and Qwen-VL \cite{bai2023qwen}. MLLMs inherit the robust linguistic and reasoning capacities of LLMs while integrating the capability to process and align information across diverse modalities (images, audio, video) within a unified, end-to-end framework \cite{li2024surveying}. This offers a paradigm-shifting prospect for affective computing. As depicted in Figure \ref{fig:MLLM end-to-end}, MLLMs can concurrently receive raw audio, video, and text data as input, implicitly acquire complex cross-modal interactions through supervised fine-tuning, and conduct end-to-end affective computing by utilizing their powerful contextual learning and instruction-following abilities.

Despite the immense potential of MLLMs in MAC, significant challenges persist in their practical application. Current MLLMs demonstrate substantial performance variability across complex MAC tasks, which can be attributed to differences in architectural designs, pre-training objectives, data scales, and inherent capabilities \cite{yin2024survey}. However, the precise relationship between these design choices and their specific impacts on MAC performance remains insufficiently explored and understood. Existing benchmarks have primarily focused on assessing unimodal LLMs or dual-modal MLLMs (e.g., text+vision, text+audio) in relatively simple tasks such as sentiment analysis \cite{chen2024emotionqueen, gao2025eemo, zhang2025can}. A critical void exists: the lack of a systematic, comprehensive evaluation of state-of-the-art MLLMs capable of processing all relevant modalities (text, audio, visual) on established MAC datasets. Such a benchmark is essential to identify which models excel at specific aspects of affective understanding, thereby guiding model selection and future development.

Furthermore, the performance of MLLMs demonstrates a marked sensitivity to the framing of tasks within prompts \cite{mohanty2025future}. Although supervised fine-tuning (SFT) has been empirically validated to boost their task-specific efficacy, the potential of advanced prompt engineering strategies to unlock and optimize their inherent affective computing capabilities remains substantially underexplored in current research.

To address these critical gaps, we carry out a thorough benchmark assessment of open-source MLLMs capable of processing audio, visual, and textual modalities concurrently. Our evaluation spans multiple well-established MAC datasets, including CMU-MOSI \cite{zadeh2016mosi}, CMU-MOSEI \cite{yu2020ch}, CH-SIMS \cite{yu2020ch}, CH-SIMS v2 \cite{liu2022make}, MELD \cite{poria2018meld}, and UR-FUNNY v2 \cite{hasan2019ur}. This evaluation not only compares MLLMs against each other but also juxtaposes their performance with traditional machine learning methods to quantify the advancements and identify remaining challenges. Additionally, we perform an in-depth analysis to elucidate how model architecture characteristics (e.g., modality alignment mechanisms, fusion strategies, model size) and dataset properties (e.g., modality dominance, domain) influence performance in affective analysis.

To enhance the performance of MLLMs in MAC, we propose a simple but effective strategy that integrates generative knowledge prompting \cite{liu2021generated} with SFT. Specifically, we first leverage the zero-shot capability of MLLMs to extract descriptions from both audio and video modalities. Subsequently, we design knowledge-guided prompts to effectively incorporate these extracted cues into the model input, followed by SFT on the augmented input. Experimental results validate that this strategy outperforms standalone SFT methods, achieving significant improvements in MLLM performance across affective computing tasks.

The main contributions are summarized as follows:
\begin{itemize}
    \item We conduct the first systematic evaluation of state-of-the-art MLLMs capable of simultaneous processing of audio, visual, and textual modalities. 
    \item We reveal the mechanisms by which model architectural designs and dataset characteristics influence MLLMs' performance in affective analysis tasks, providing actionable insights for model optimization.
    \item We propose a hybrid strategy that integrates generative knowledge prompting with SFT. Experimental results demonstrate that this approach significantly enhances MLLMs' performance in affective computing tasks.
\end{itemize}

\section{Related Work}

\subsection{Multimodal Affective Computing}
MAC seeks to recognize and analyze human emotions by integrating information from multiple modalities. Traditional methods often rely on early fusion \cite{hazarika2018self, keshari2019emotion}, late fusion \cite{song2018decision, pandeya2021deep}, or attention-based strategies \cite{praveen2022joint, zou2022speech}. Although these techniques outperform unimodal methods, they still fail to adequately capture the complex cross-modal interplay of affective cues \cite{sanku2024effective}. This limitation has spurred a recent shift towards MLLMs \cite{yang2025omni, murzaku2025omnivox}. MLLMs leverage their unified semantic space and emergent reasoning ability to more effectively detect subtle interactions between modalities and understand emotions within specific contexts \cite{zhang2025mellm}. 

\subsection{Multimodal Large Language Models}
MLLMs are built upon LLMs and integrate multimodal encoders \cite{radford2021learning, tong2022videomae, chen2022beats} through projection or cross-attention to enable unified multimodal processing. In recent years, the open-source community has yielded powerful MLLMs, ranging from vision-language models with advanced visual reasoning capabilities (e.g., BLIP-2 \cite{li2023blip}, LLaVA \cite{liu2023visual}) to audio-language models exhibiting robust audio comprehension (e.g., SALMONN \cite{tang2023salmonn}, Qwen-Audio \cite{chu2023qwen}). More recently, the development of MLLMs is progressing towards comprehensive omnimodal models that unify multiple modalities within a single framework \cite{yao2024minicpm, zhao2025humanomni, liu2025ola}. For instance, Qwen2.5-Omni \cite{xu2025qwen2} perceives diverse modalities, including text, images, audio, and video, while simultaneously generating text and natural speech responses in a streaming manner. These omnimodal models not only capture complex relationships between text, vision, and sound but also demonstrate enhanced robustness in real-world scenarios. In this study, we selected multiple open-source MLLMs that support joint modeling of text, video, and audio for benchmarking purposes.

\begin{table*}[t]
    \caption{Comparison of MLLMs and their components.}
    \centering
    \begin{tabular}{lccc}
        \toprule[2pt]
        MLLM&Visual Encoder&Audio Encoder&LLM\\
        \midrule[1pt]
        Qwen2.5Omni&Qwen2.5-VL&Whisper-large-v3&Qwen2.5 (7B)\\
        HumanOmni&SigLIP&Whisper-large-v3&Qwen2.5 (7B)\\
        Ola&SigLIP&BEATs&Qwen2.5 (7B)\\
        VideoLLaMA2-AV&CLIP&BEATs&Qwen2 (7B)\\
        MiniCPM-o&SigLip&Whisper-medium&Qwen2.5 (7B)\\
        PandaGPT&ImageBind&ImageBind&Vicuna (7B)\\
        Emotion-LLaMA&MAE,VideoMAE,EVA&HuBERT&llama2 (7B)\\
        \bottomrule[2pt]
    \end{tabular}
    \label{tab:MLLM_description}
\end{table*}

\subsection{Prompting Strategy}
Effective prompting strategies are crucial for enhancing MLLMs’ affective reasoning capabilities. Recent research demonstrates their potential for MAC. For instance, methods like Multi-Views Prompt Learning \cite{xu2024learning} effectively capture the emotional cues involved in different levels of semantic information, while Set-of-Vision-Text Prompting (SoVTP) \cite{wang2025visual} preserves holistic scene context by overlaying spatial annotations on full-scene inputs and integrating auxiliary cues like body posture, environment, and social dynamics. Additionally, combining prompts with acoustic analysis or Chain-of-Thought (CoT) reasoning has shown promise in emotion recognition in conversation tasks \cite{murzaku2025omnivox}.

However, existing work has primarily explored bimodal scenarios in MAC tasks. To address this gap in trimodal tasks, we propose a strategy combining generative knowledge prompting across text, audio, and video modalities.

\section{Benchmark}
\subsection{Datasets}

In this study, we employ six datasets, encompassing multimodal sentiment analysis (MSA) datasets (CMU-MOSI \cite{zadeh2016mosi}, CMU-MOSEI \cite{zadeh2018multimodal}, CH-SIMS \cite{yu2020ch}, and CH-SIMS v2 \cite{liu2022make}), multimodal emotion recognition (MER) dataset (MELD \cite{poria2018meld}), and multimodal humor detection (MHD) dataset (UR-FUNNY v2 \cite{hasan2019ur}). Here, We present a concise overview of these datasets below, with detailed statistics summarized in the Appendix.

\subsubsection{CMU-MOSI and CMU-MOSEI}
The CMU-MOSI dataset \cite{zadeh2016mosi} consists of 93 YouTube videos, which are divided into 2,199 clips, with each clip annotated with sentiment scores on a 7-point scale ranging from strong negative (-3) to strong positive (+3). Likewise, the CMU-MOSEI dataset \cite{zadeh2018multimodal} encompasses 23,453 video clips derived from various online platforms and adheres to the same sentiment score labeling scheme.
\subsubsection{CH-SIMS and CH-SIMS v2}
The CH-SIMS dataset \cite{yu2020ch} contains 2,281 refined video segments sourced from movies, TV series, and variety shows, with sentiment annotations ranging from negative (-1) to positive (+1) for each clip. The CH-SIMS v2 dataset \cite{liu2022make} extends this corpus to 4,402 supervised segments and 10,161 unsupervised segments (totaling 14,563 clips), collected from 11 diverse scenarios like vlogs, interviews, and talk shows, emphasizing richer non-verbal behaviors while retaining the original annotation methodology.
\subsubsection{MELD}
The MELD dataset \cite{poria2018meld} is a multimodal corpus specifically designed for emotion recognition in conversational contexts. This dataset is constructed based on dialogues from the television series "Friends", comprising over 1,400 conversational sequences containing 13,000 speaker utterances. Each utterance is annotated with one of the seven basic emotional categories (anger, disgust, sadness, joy, neutral, surprise, fear) as well as sentiment polarity labels (positive, negative, neutral).
\subsubsection{UR-FUNNY v2} The UR-FUNNY v2 dataset \cite{hasan2019ur} is a diverse multimodal resource for humor detection in natural language processing. Compared with the original UR-FUNNY dataset, it removes noisy and overlapping instances from the original dataset. In terms of content composition, UR-FUNNY v2 incorporates a greater number of contextual sentences compared to its predecessor, which enriches the contextual information available for analysis.

\subsection{Multimodal Large Language Models}

To achieve end-to-end affective computing, the evaluated MLLMs must support the collaborative input of audio, video, and text. Additionally, the models must be open-source to enable effective SFT. Based on the above requirements, this study selects HumanOmni \cite{zhao2025humanomni}, Qwen2.5Omni \cite{xu2025qwen2}, VideoLLaMA2-AV \cite{cheng2024videollama}, Ola \cite{liu2025ola}, MiniCPM-o 2.6 \cite{yao2024minicpm}, Emotion-LLaMA \cite{cheng2024emotion}, and PandaGPT \cite{su2023pandagpt} as the experimental models. Their basic information is summarized in Table \ref{tab:MLLM_description}, and detailed characteristics can be found in Appendix A.

\begin{figure*}[t]
    \centering
    \includegraphics[width=1\textwidth]{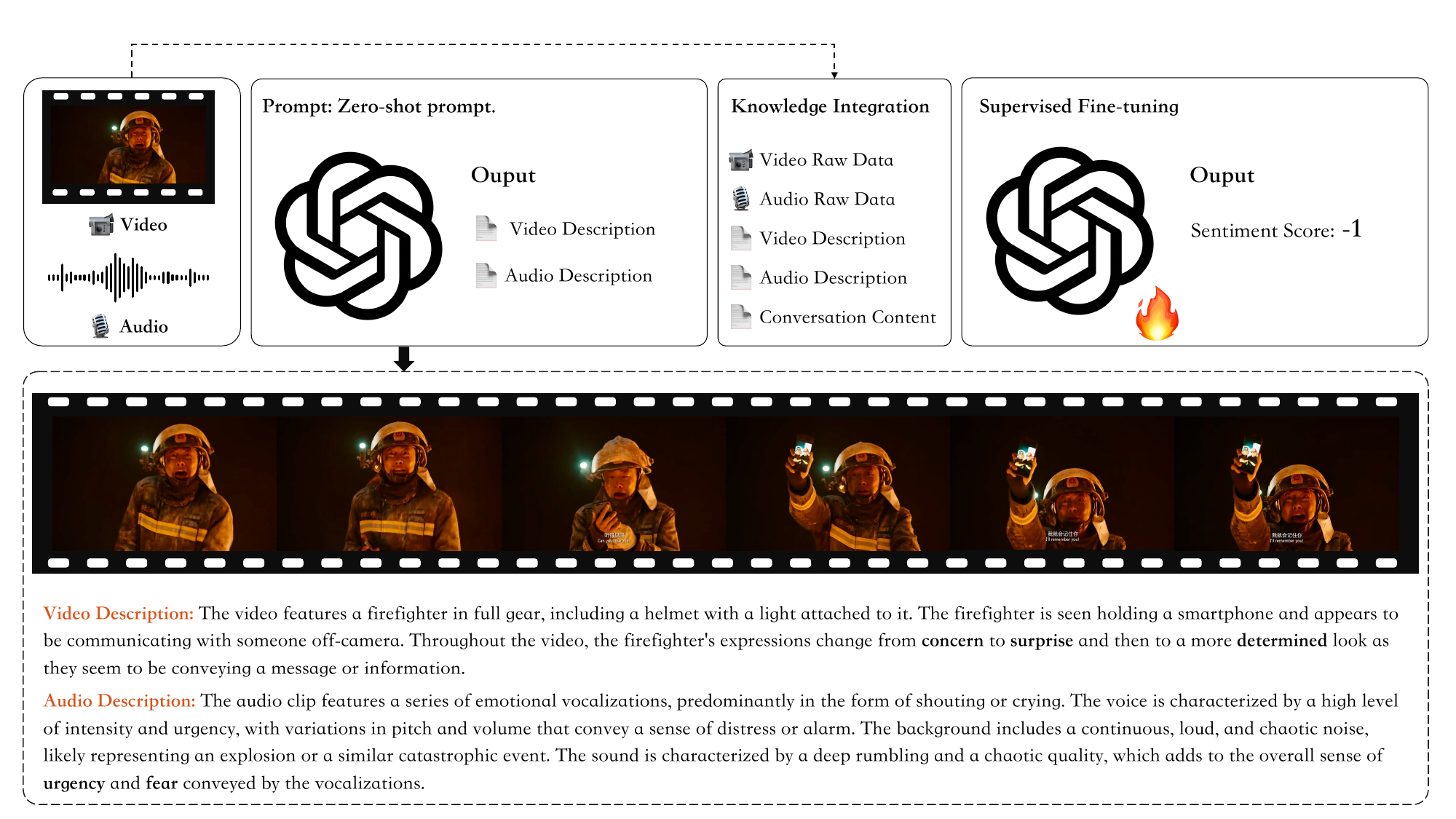} 
    \caption{Enhancing MLLM Performance in MAC via knowledge generation and supervised fine-tuning.}
    \label{fig:prompt strategy}
\end{figure*}
\subsection{Method Overview}
\subsubsection{Supervised Fine-tuning}
To enhance the adaptability of MLLMs for MAC tasks like MSA, MER, and MHD, the employment of SFT to adjust model parameters is adopted. As a task-specific optimization paradigm built on pre-trained models, SFT leverages labeled datasets—comprising input samples and their corresponding target outputs—to further refine model parameters, thereby enabling the model to achieve better alignment with the characteristics and requirements of specific downstream tasks.

Furthermore, to mitigate computational overhead, we incorporated Low-Rank Adaptation (LoRA) technology \cite{hu2022lora}. Instead of directly modifying all model parameters, LoRA implements fine-tuning by injecting low-rank matrices into the model's weight matrices. Specifically, in LoRA-based fine-tuning, two low-rank matrices $A$ and $B$ are introduced. A rank-$r$ matrix $\Delta W=A \times B$ is then constructed from these two matrices and added to the original weight matrix $W$. The formula is as follows:

\begin{equation}
W_{new}=W+A\times B
\end{equation}

Here, $W$ is the original weight matrix of the pre-trained model, which is typically kept fixed during fine-tuning. $A$ and $B$ are the low-rank matrices that need to be trained, and the number of parameters in these matrices is significantly smaller than that of the original weight matrix $W$.

\subsubsection{Prompt Strategy}
To enhance the performance of MLLMs in MAC, we propose an innovative strategy that synergistically integrates generative knowledge prompting with SFT. As illustrated in Figure \ref{fig:prompt strategy}, our approach commences by leveraging the zero-shot capabilities of MLLMs to extract salient descriptions pertaining to affective computing directly from raw video and audio inputs. This initial step is followed by a comprehensive knowledge aggregation phase, wherein we systematically consolidate multimodal data streams—including original audiovisual data, their corresponding generated descriptions, and textual dialogue content—into a unified input framework for the MLLM. Subsequent SFT is designed to align the model's output distribution with the specific requirements of MAC tasks, thereby ensuring optimal adaptation to the nuances of affective analysis. By focusing on getting emotion-related semantic features from unstructured multimedia data, this approach helps the MLLM pay more refined attention to affective cues.

\section{Experiments}

We conducted experiments using multiple MAC datasets and several open-source MLLMs. Due to space constraints, detailed information on evaluation metrics and model details is provided in the Appendix.

\begin{table*}[t]
    \caption{The comparison with baselines on the CMU-MOSI and CMU-MOSEI dataset. The results of the baselines denoted with † are directly sourced from their respective publications. The best results are in bold, and the runner-up results are underlined.}
    \centering
    \begin{tabular}{l|ccccc|ccccc}
    \toprule[2pt]
    \multirow{2}{*}{Models} & \multicolumn{5}{c|}{CMU-MOSI} & \multicolumn{5}{c}{CMU-MOSEI} \\ 
     & Acc7↑ & Acc2↑ & F1↑ & MAE↓ & Corr↑ & Acc7↑ & Acc2↑ & F1↑ & MAE↓ & Corr↑ \\ 
    \midrule[2pt] 
    C-MIB† & 48.2 & 85.2 & 85.2 & 0.728 & 0.793 & 53.0 & 86.2 & 86.2 & 0.584 & 0.789 \\
    MGT† & 50.4 & 86.3 & 86.3 & 0.659 & 0.822 & 54.3 & 86.1 & 86.1 & 0.522 & 0.764 \\
    KAN-MCP† & 48.3 & 89.4 & 89.4 & 0.615 & 0.857 & 53.9 & \textbf{87.7} & \textbf{87.6} & 0.522 & 0.788 \\
    MOAC† & 48.6 & 89.0 & 89.0 & 0.605 & 0.857 & 54.3 & \underline{87.6} & \textbf{87.6} & 0.512 & 0.793  \\ 
    PandaGPT & 52.1 & 90.2 & 90.2 & 0.536 & \textbf{0.899} & 54.6 & 87.3 & \underline{87.1} & 0.628 & 0.800  \\ 
    Emotion-LLaMA & 40.7 & 86.1 & 86.2 & 0.800 & 0.764 & 51.9 & 83.7 & 82.6 & 0.704 & 0.751  \\
    MiniCPM-o & 49.8 & 89.5 & 89.5 & 0.636 & 0.853 & 51.2 & 86.6 & 86.3 & 0.553 & 0.766  \\
    Ola & 48.3 & 89.3 & 89.3 & 0.620 & 0.860 & 54.3 & 84.4 & 83.5 & 0.534 & 0.778  \\ 
    VideoLLaMA2-AV & 50.4 & 90.5 & 90.5 & 0.571 & 0.877 & 57.9 & 84.2 & 83.2 & 0.493 & 0.802  \\  
    Qwen2.5Omni & \underline{53.9} & 90.5 & 90.5 & \underline{0.523} & \textbf{0.899} & 53.2 & 80.0 & 78.0 & 0.563 & 0.730  \\
    HumanOmni & 52.8 & \textbf{91.3} & \textbf{91.3} & 0.549 & 0.881 & \underline{58.6} & 86.1 & 85.4 & \underline{0.483} & \underline{0.807}  \\ 
    \midrule[1pt]
    HumanOmni(Optimized) & \textbf{55.9} & \underline{90.8} & \underline{90.8} & \textbf{0.510} & \underline{0.896} & \textbf{58.7} & 86.3 & 85.6 & \textbf{0.478} & \textbf{0.810}  \\   
    \bottomrule[2pt]
    \end{tabular}
    \label{tab:cmu_result}
\end{table*}

\begin{table*}[t]
    \caption{The comparison with baselines on the CH-SIMS and CH-SIMS v2 dataset. The results of the baselines denoted with † are directly sourced from their respective publications. The best results are in bold, and the runner-up results are underlined.}
    \centering
    \begin{tabular}{l|ccccc|ccccc}
    \toprule[2pt]
    \multirow{2}{*}{Models} & \multicolumn{5}{c|}{CH-SIMS} & \multicolumn{5}{c}{CH-SIMS v2} \\ 
     & Acc5↑ & Acc2↑ & F1↑ & MAE↓ & Corr↑ & Acc5↑ & Acc2↑ & F1↑ & MAE↓ & Corr↑ \\ 
    \midrule[2pt]
    HGTFM† & 44.0 & 80.5 & 80.3 & 0.410 & 0.598 & 58.0 &  82.9 & 82.9 & 0.279 & 0.740 \\
    KAN-MCP† & - & - & - & - & - & 57.3 & 81.6 & 81.7 & 0.281 & 0.742\\
    PandaGPT & 38.3 & 77.2 & 74.7 & 0.431 & 0.537 & 46.2 & 72.3 & 72.0 & 0.378 & 0.557  \\ 
    Emotion-LLaMA & 41.1 & 77.2 & 75.4 & 0.403 & 0.628 & 37.9 & 74.7 & 73.5 & 0.359 & 0.632 \\ 
    MiniCPM-o & 48.8 & 82.5 & 80.5 & 0.350 & 0.695 & 56.3 & 83.8 & 83.7 & 0.267 & 0.748  \\ 
    Ola & 48.4 & 81.6 & 80.2 & 0.406 & 0.646 & 59.5 & 81.1 & 81.2 & 0.309 & 0.685  \\ 
    VideoLLaMA2-AV & \underline{52.1} & 81.6 & 82.3 & 0.388 & 0.733 & 40.5 & 83.7 & 83.8 & 0.382 & 0.750  \\ 
    Qwen2.5Omni & 46.8 & 82.3 & 80.1 & \underline{0.310} & \underline{0.758} & 61.7 & \textbf{86.9} & \textbf{86.9} & \textbf{0.211} & \textbf{0.841} \\ 
    HumanOmni & \underline{52.1} & \underline{85.1} & \underline{85.0} & 0.327 & 0.749 & \underline{62.8} & 85.9 & 85.9 & 0.266 & 0.795 \\ 
    \midrule[1pt]
    HumanOmni(Optimized) & \textbf{59.1} & \textbf{86.0} & \textbf{86.3} & \textbf{0.294} & \textbf{0.770} & \textbf{63.2} & \underline{86.1} & \underline{86.0} & \underline{0.249} & \underline{0.804} \\ 
    \bottomrule[2pt]
    \end{tabular}
    \label{tab:sims_result}
\end{table*}

\subsection{Evaluation Baselines}
In this study, we employed the MLLMs, previously introduced, as the baseline models and further conducted a comparative analysis with state-of-the-art (SOTA) multimodal machine learning (MML) methods.
\subsubsection{MLLM}
As shown in Table \ref{tab:MLLM_description}, the MLLMs we selected include Qwen2.5Omni, HumanOmni, Ola, VideoLLaMA2-AV, MiniCPM-o, PandaGPT, and Emotion-LLaMA. Among them, Qwen2.5Omni, HumanOmni, Ola, and MiniCPM-o are based on the same large language model (LLM), namely Qwen2.5 (7B); VideoLLaMA2-AV is based on Qwen2 (7B), PandaGPT is based on Vicuna (7B), and Emotion-LLaMA is based on LLaMA2 (7B).
\subsubsection{MML} For comparative analysis, we selected the SOTA methods for each dataset. Specifically, or the CMU-MOSI and CMU-MOSEI datasets, we chose MOAC \cite{mai2025learning}, C-MIB\cite{mai2022multimodal}, MGT\cite{mai2025injecting}, and KAN-MCP\cite{luo2025towards} as baseline methods. For the CH-SIMS dataset, we used HGTFM \cite{yang2025hgtfm} as the primary comparison benchmark. For the CH-SIMS v2 dataset, we selected HGTFM and KAN-MCP as comparison benchmarks. For the UR-FUNNY v2 dataset, we adopted SemanticMAC \cite{lin2024end} as the reference method. For the MELD dataset, we used SemanticMAC and MGT as comparison baselines.

\begin{table*}[t]
    \caption{The comparison with baselines on the UR-FUNNY v2 and MELD dataset. The results of the baselines denoted with † are directly sourced from their respective publications. The best results are in bold, and the runner-up results are underlined.}
    \centering
    \begin{tabular}{l|cccc|cc}
    \toprule[2pt]
    \multirow{2}{*}{Models} & \multicolumn{4}{c|}{UR-FUNNY v2} & \multicolumn{2}{c}{MELD} \\ 
    & w-Precision ↑ & w-Recall ↑ & w-Acc ↑ & w-F1 ↑  & w-Acc↑ & w-F1↑ \\ 
    \midrule[2pt]
    SemanticMAC† & 76.1 & 75.6 & 75.6 & 75.5 & 62.2 & 61.4 \\
    MGT† & - & - & - & - & 65.8 & 63.8 \\
    PandaGPT  & 75.0 & 74.7 & 74.7 & 74.7 & 63.4 & 62.4 \\ 
    Emotion-LLaMA  & 72.8 & 72.3 & 72.3 & 72.2 & 63.2 & 60.0  \\ 
    MiniCPM-o & 76.5 & 75.6 & 75.6 & 75.4 & 65.8 & 62.1 \\ 
    Ola & \textbf{80.9} & \textbf{80.9} & \textbf{80.9} & \textbf{80.9} & 62.9 & 56.9 \\ 
    VideoLLaMA2-AV & 76.0 & 71.4 & 71.4 & 70.2 & 67.8 & 66.2 \\ 
    Qwen2.5Omni & 70.9 & 61.2 & 61.2 & 55.7 & 66.6 & 64.9 \\ 
    HumanOmni & 78.6 & 78.3 & 78.3 & 78.2 & \underline{68.9} & \underline{66.6} \\ 
    \midrule[1pt]
    HumanOmni(Optimized) & \underline{79.9} & \underline{79.9} & \underline{79.9} & \underline{79.9} & \textbf{69.0} & \textbf{67.2} \\ 
    \bottomrule[2pt]
    \end{tabular}
    \label{tab:meld}
\end{table*}

\subsection{Supervised Fine-tuning Details}
To adapt MLLMs to the task of MAC, we conducted supervised fine-tuning on MLLMs across six different datasets. Specifically, Qwen2.5Omni and MiniCPM-o models were fine-tuned using the LLaMA-Factory framework \cite{zheng2024llamafactory}, while the other models were fine-tuned using the code from their respective open-source repositories. During the fine-tuning process, we incorporated the FlashAttention-2 \cite{dao2023flashattention} to optimize the attention module of transformers, effectively reducing memory consumption and computational time. Additionally, to further reduce computational costs, we employed BF16 precision and utilized the DeepSpeed library to achieve distributed training.

In terms of hyperparameter settings, the training epoch for Emotion-LLaMA was selected from {10,20,30,40}, while that for PandaGPT was chosen within the range of 1 to 10. For the remaining models, the training epoch was selected from {1,2,3}. The learning rate of the models was adjusted within the range of 1e-6 to 1e-3. For the LoRA module, the rank and $\alpha$ parameters were set to {8, 16, 64, 128, 256} and {16, 32, 128, 256, 512}, respectively. During fine-tuning, we monitored the model's accuracy on the validation set to select the optimal inference checkpoint. All experiments were conducted on four NVIDIA RTX 4090 48G GPUs. 

Regarding model training strategies, VideoLLaMA2-AV, Ola, and HumanOmni adopted a two-stage training and fine-tuning approach. In the first stage, the LLM parameters were frozen, focusing on training the audio and visual encoders and projectors to enable the model to efficiently extract and understand audio and visual information. In the second stage, the parameters of the audio and visual encoders and projectors were frozen, and the LLM was fine-tuned using the efficient LoRA fine-tuning technique to achieve a deep integration of visual, audio features, and language information, thereby further enhancing the model's performance in multimodal affective computing tasks.

In contrast, Qwen2.5Omni, MiniCPM-o, PandaGPT, and Emotion-LLaMA employed a single training strategy based on LoRA fine-tuning, directly optimizing the language model to adapt to specific task requirements.

\section{Results and Discussion}

\subsection{Main Results}
\subsubsection{Results on MSA}
As shown in Table \ref{tab:cmu_result}, MLLMs demonstrate exceptional performance on the CMU-MOSI dataset. This outstanding performance can be attributed to the dominant role of the text modality in this dataset \cite{mai2019divide}—MLLMs can fully leverage their robust language understanding and generation capabilities by fine-tuning the language model module, thereby achieving significant performance improvements in relevant tasks and ultimately yielding excellent results on the CMU-MOSI dataset. Specifically, in the testing phase of this dataset, except for Emotion-LLaMA, all other MLLMs significantly outperform MOAC across most evaluation metrics. Among them, Qwen2.5Omni leads by 5.3\% in the Acc7 metric, and HumanOmni is 2.3\% higher in the Acc2 metric, with particularly notable advantages.

However, the performance of MLLMs on the CMU-MOSEI dataset shows a divergent trend. In terms of the Acc7 metric, only HumanOmni and VideoLLaMA2-AV outperform MOAC, with improvements of 4.3\% and 3.6\% respectively, while the remaining MLLMs lag behind this MML model in multiple metrics. Regarding the Acc2 metric, all MLLMs perform worse than MOAC. An analysis of the training data distribution and model output results of the CMU-MOSEI dataset reveals that the dataset has a significant issue of sample distribution imbalance—21.7\% of the samples are labeled as 0. This imbalance directly causes the fine-tuned MLLMs to tend to output 0 labels during prediction, ultimately leading to poor performance in metrics such as Acc2. This phenomenon profoundly highlights the significant impact of dataset sample distribution on the sentiment analysis performance of MLLMs, suggesting that data balance is a key factor to be prioritized in model optimization.

\begin{figure}[t]
    \centering
    \includegraphics[width=0.48\textwidth]{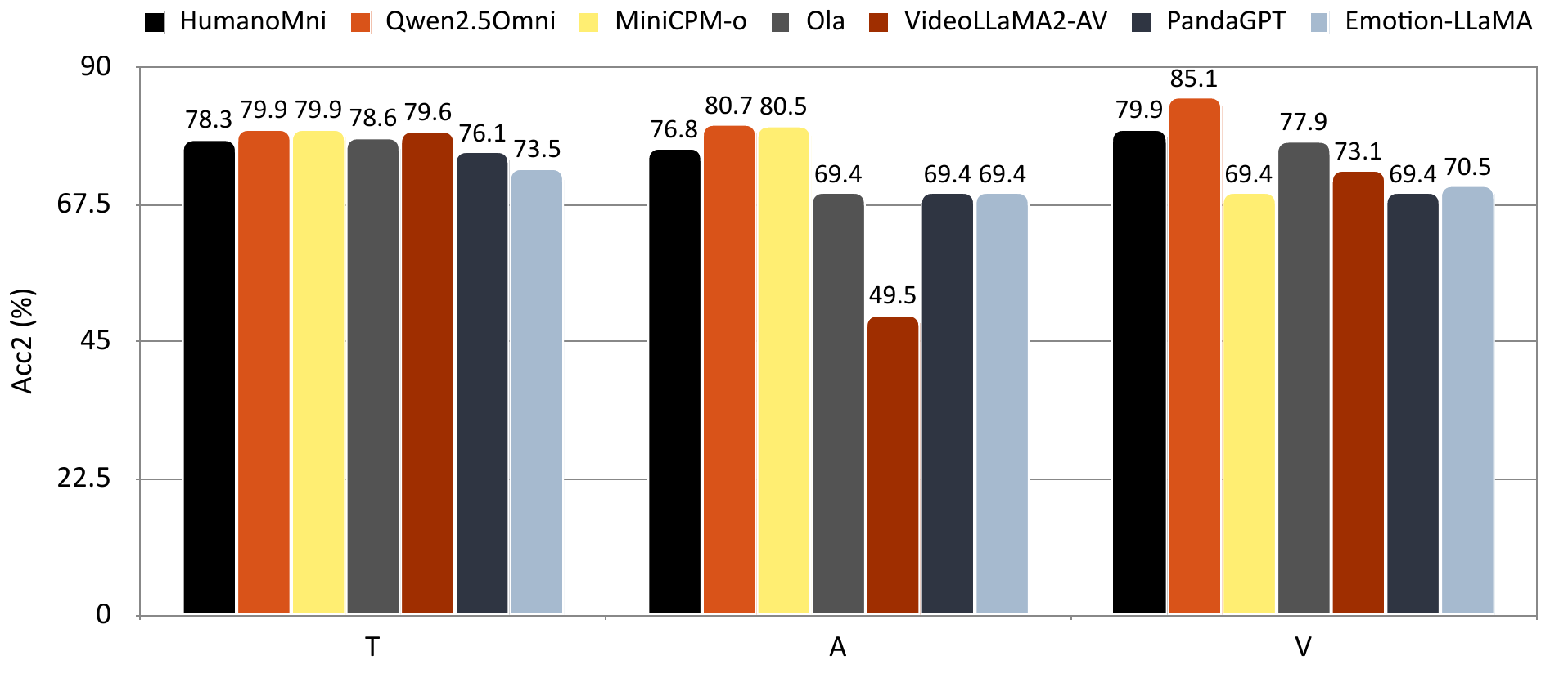} 
    \caption{Performance comparison of MLLMs in unimodal settings on the CH-SIMS dataset.}
    \label{fig:sims-unimodal}
\end{figure}

\begin{figure}[t]
    \centering
    \includegraphics[width=0.48\textwidth]{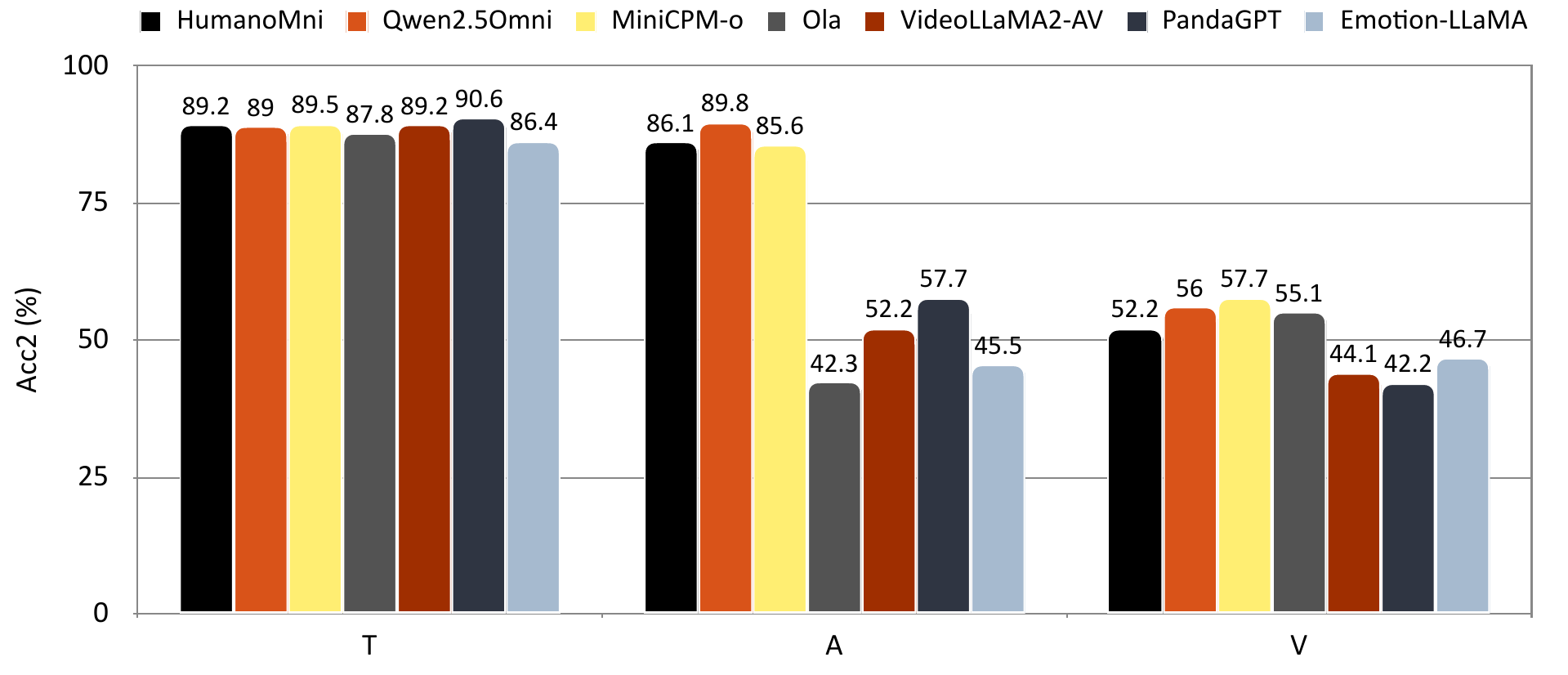} 
    \caption{Performance comparison of MLLMs in unimodal settings on the CMU-MOSI dataset.}
    \label{fig:mosi-unimodal}
\end{figure}

As shown in Table \ref{tab:sims_result}, in CH-SIMS and CH-SIMS v2, all MLLMs achieve excellent performance except for PandaGPT and Emotion-LLaMA, which perform relatively poorly. In the CH-SIMS dataset, HumanOmni performs the best; compared with HGTFM, its Acc5 is improved by 8.1\%, Acc2 by 4.6\%, and F1 by 4.7\%. In CH-SIMS v2, Qwen2.5Omni is the optimal model, with Acc5 improved by 3.7\%, Acc2 and F1 both improved by 4.0\% compared with HGTFM. The above results indicate that MLLMs exhibit more prominent performance advantages in datasets where the contributions of various modalities are more balanced \cite{yu2020ch}, which further verifies their strong ability in fusing and processing multimodal information. Especially in data environments with good modal synergy, they can better exert their architectural advantages.

\subsubsection{Results on MER and MHD}
In the task of MER, MLLMs all demonstrate excellent performance on the MELD dataset. As shown in Table \ref{tab:meld}, compared with SemanticMAC, the HumanOmni model achieves a 6.7\% improvement in w-Acc and a 5.2\% improvement in w-F1.

In the task of MHD, the experimental results on the UR-FUNNY v2 dataset are presented in Table \ref{tab:meld}. The Ola, MiniCPM-o, and HumanOmni models perform better than SemanticMAC, while the performance of the remaining MLLMs is inferior to this benchmark model. Among these better-performing models, the Ola model stands out with its w-Acc being 4.8\% higher than that of SemanticMAC. Notably, although the Ola model shows average performance on multiple datasets, it exhibits excellent performance on the UR-FUNNY v2 dataset. This phenomenon indicates that different MLLMs have significant differences in their adaptability to specific datasets.

\subsection{Enhancing MLLM with Prompt Engineering}
To enhance the MAC capability of MLLMs, we optimized the high-performing HumanOmni model by adopting a combined strategy of generative knowledge prompting and supervised fine-tuning. The results are presented in Table \ref{tab:cmu_result}, \ref{tab:sims_result} and \ref{tab:meld}. This strategy outperformed the original simple fine-tuning on all datasets, with particularly significant improvements in multi-class accuracy (Acc). Specifically, the Acc5 on CH-SIMS increased by 7.0\%, and the Acc7 on CMU-MOSI rose by 3.1\%. This indicates that supplementing the model with descriptive knowledge of audio and video can strengthen its understanding of the deep correlations between multimodal emotional features, thereby improving classification accuracy in complex scenarios.

However, in CMU-MOSEI and CH-SIMS v2, the improvement from this strategy was marginal. This may be because the emotional features in these two datasets are relatively distinct, and simple SFT alone enables the model to sufficiently learn the core discriminative information. In such cases, the additional descriptive information fails to provide effective gains and may even slightly interfere with the model's judgments due to information redundancy.

\subsection{Case Study}
The workflow of our prompting strategy in affective computing is illustrated in Figure \ref{fig:prompt strategy}. Firstly, raw video and audio data are input into the model, which performs precise analysis to identify key emotional clues and generate descriptions. For the video, the model captures visual information such as ``a helmet with a light attached to it. The firefighter is seen holding a smartphone and appears to be communicating with someone off-camera'', while also meticulously recording the emotional changes: ``the firefighter's expressions change from concern to surprise and then to a more determined''. In the audio analysis, the model identifies sound elements like `shouting', `crying', and `continuous, noisy, and loud noises', and matches them with emotional information such as `pain', `alarm', `fear', and `sense of urgency'.

After the initial extraction of key clue descriptions, we re-input the raw video, audio data, generated descriptions, and dialogue texts into the model for in-depth reasoning. By fully integrating this multimodal information, the model ultimately outputs an emotional score of -1 (representing extremely negative emotion). Through the extraction of multimodal emotional description information, the model can more accurately grasp the emotional context in videos and audios, and the final experimental results verify the effectiveness of this prompting strategy.

\subsection{Analysis of Input Impact}

To investigate the contribution mechanisms of different modalities in MLLMs, this study selected the HumanOmni model, which demonstrates excellent performance across multiple datasets, to conduct unimodal analysis experiment. 

As shown in Figure \ref{fig:sims-unimodal}, on the CH-SIMS dataset, the text modality exhibits a common advantage—all MLLMs achieve superior performance, indicating that the current mechanisms for processing textual information in models possess cross-model universal effectiveness. In terms of the audio modality, the prediction performance of VideoLLaMA2-AV is significantly lower than the average level, revealing that this model may have design limitations in aspects such as audio feature encoding, the mapping of acoustic information to the semantic space, or cross-modal alignment mechanisms, making it difficult to effectively capture key information in the audio modality. In sharp contrast, HumanOmni, Qwen2.5Omni, and MiniCPM-o perform prominently in the audio modality, suggesting that these three models possess more robust modality modeling capabilities in the audio signal processing pipeline. Regarding the visual modality, Qwen2.5Omni outperforms other comparative models by a significant margin. This result indicates that the model has notable technical advantages in the visual feature extraction stage, and its visual encoder and modality fusion mechanism can better adapt to the characteristics of visual tasks in the CH-SIMS dataset, thereby more accurately capturing key visual information in video frames and converting it into effective semantic representations.

As shown in Figure \ref{fig:mosi-unimodal}, on the CMU-MOSI dataset, where the text modality dominates, the text modality also shows a consistent advantage, with all MLLMs maintaining excellent performance. In the audio modality, the test results of HumanOmni, Qwen2.5Omni, and MiniCPM-o are significantly better than those of other MLLMs. In-depth analysis reveals that all three models employ Whisper as the audio encoder, and this encoder has undergone sufficient training for speech-to-text tasks during the pre-training phase. This technical characteristic enables its performance in the standalone audio modality to be comparable to that of the text modality. The above results confirm a key conclusion: the degree of adaptation between the pre-training tasks of the audio encoder and downstream sentiment analysis tasks directly affects the performance of the model.

\section{Conclusions}

In this paper, we have systematically evaluated state-of-the-art MLLMs capable of simultaneous processing of audio, visual, and textual modalities. Our comprehensive benchmark assessment across multiple MAC datasets has revealed how model architectural designs and dataset characteristics influence MLLMs’ performance in affective analysis tasks. We have also proposed a hybrid strategy that integrates generative knowledge prompting with supervised fine-tuning, which has significantly enhanced MLLMs’ performance in affective computing tasks. These findings offer valuable insights for model optimization and highlight the potential of advanced prompt engineering strategies in unlocking the full capabilities of MLLMs for affective computing. Future work can further explore the optimization of MLLMs in more complex and diverse MAC scenarios, as well as the development of more sophisticated prompting strategies to continue pushing the boundaries of affective computing technology.

\bibliographystyle{IEEEtran}
\bibliography{custom}

\appendix
\subsection{Evaluation Metrics}
We use different evaluation metrics for different datasets based on their label types:
\subsubsection{CMU-MOSI and CMU-MOSEI datasets}
We use the following evaluation metrics to measure the performance of the model: (1) Acc7: the accuracy of classifying sentiment scores into seven discrete classes (predictions rounded to nearest integer in [-3, 3]); (2) Acc2: the accuracy for positive or negative binary classes (neutral utterances excluded); (3) F1 score: the harmonic mean of precision and recall, used to evaluate performance in binary sentiment classification (neutral utterances excluded); (4) MAE: the mean absolute error between the model's predictions and the annotated sentiment labels; (5) Corr: the correlation coefficient indicating the strength and direction of the relationship between the model's predictions and human annotations. 
\subsubsection{CH-SIMS and CH-SIMS v2 datasets.}
We use the following metrics: (1) Acc5: the accuracy of dividing emotional scores into five discrete categories (predictions rounded to nearest integer in [-1, 1]); (2) Acc3: the accuracy in categorizing emotions into three types (positive, neutral, and negative); (3) Acc2, F1 score, MAE, and Corr: their meanings are the same as those of the CMU-MOSI and CMU-MOSEI dataset.
\subsubsection{MELD dataset}
We use the following metrics: (1) w-Acc: the weighted accuracy; (2) w-F1: the weighted average F1 score.
\subsubsection{UR-FUNNY v2 dataset}
We use the following metrics: (1) w-Precision: the weighted precision; (2) w-Recall: the weighted recall; (3) w-Acc; (4) w-F1. 

\subsection{Multimodal Large Language Models}
\subsubsection{Qwen2.5Omni}
The Qwen2.5-Omni \cite{xu2025qwen2} is an end-to-end multimodal model capable of processing a variety of modalities, including text, image, audio, and video, while simultaneously generating text and natural speech responses. Its core architecture follows the Thinker-Talker design. The Thinker is tasked with processing and interpreting text, audio, and video inputs to generate high-level representations and corresponding text. The Talker then streams speech tokens based on the high-level representations created by the Thinker. This architecture enables Qwen2.5-Omni to achieve efficient pre-filling, real-time multimodal understanding, and concurrent generation of text and speech signals.

The base language model of Qwen2.5-Omni is a Transformer decoder, initialized from Qwen2.5 \cite{hui2024qwen2}. Its audio encoder is based on Whisper-large-v3 \cite{radford2023robust}, and the video encoder inherits from Qwen2.5-VL \cite{bai2025qwen2} and employs a Vision Transformer (ViT) \cite{han2022survey} based architecture. Additionally, Qwen2.5-Omni introduces TMRoPE (Time-aligned Multimodal RoPE), a novel positional encoding algorithm. By decomposing the original rotary embedding into temporal, height, and width components and applying them to different modalities respectively, TMRoPE effectively aligns the temporal information of audio and video, thereby enhancing multimodal integration.

\subsubsection{HumanOmni}
The HumanOmni \cite{zhao2025humanomni} is a large vision-speech language model designed to focus on human-centric video understanding. Its key innovation lies in the ability to simultaneously process visual and speech information in human-centric scenes. The model comprises three specialized branches for understanding face-related, body-related, and interaction-related scenes. An instruction-driven fusion module dynamically adjusts the fusion weights of features from these branches based on user instructions, enhancing the model’s flexibility and adaptability.

HumanOmni employs SigLIP \cite{zhai2023sigmoid} as visual encoders and Qwen2.5 \cite{hui2024qwen2} as base large language model. For audio processing, it uses the audio preprocessor and encoder from Whisper-large-v3 \cite{radford2023robust}, leveraging MLP2xGeLU \cite{li2024llava} to map audio features into the text domain, thus integrating them with visual and textual features.

\subsubsection{Ola}
The Ola \cite{liu2025ola} is an omnimodal language model capable of processing text, images, videos, and audio inputs, achieving competitive performance in image, video, and audio understanding tasks. Its core architecture is built upon Qwen2.5, incorporating advanced visual and audio encoding capabilities.
The visual encoder of Ola employs OryxViT \cite{liu2024oryx}, which is initialized from SigLIP-400M and preserves the original aspect ratio of images or video frames for arbitrary-resolution visual input processing. Ola introduces a Local-Global Attention Pooling layer to reduce the token length of visual features while minimizing information loss. For audio encoding, Ola adopts a dual-encoder approach, utilizing Whisper-v3 as the speech encoder and BEATs \cite{chen2022beats} as the music encoder. By concatenating the embedding features of speech and music encoders across the channel dimension, Ola achieves comprehensive audio feature extraction.

\subsubsection{VideoLLaMA2-AV}
The VideoLLaMA2 is a Video Large Language Model (Video-LLM) designed to enhance spatial-temporal modeling and audio understanding in video and audio-related tasks. Built upon its predecessor, VideoLLaMA2 introduces a tailored Spatial-Temporal Convolution (STC) connector to effectively capture the intricate spatial and temporal dynamics of video data. 

VideoLLaMA2 adopts a dual-branch framework comprising a Vision-Language Branch and an Audio-Language Branch. The language decoders are initialized with Qwen2 \cite{team2024qwen2}. The Vision-Language Branch utilizes the CLIP (ViT-L/14) model \cite{radford2021learning} as its vision backbone, processing video frames individually. The Audio-Language Branch employs BEATs, a cutting-edge audio encoder, to extract audio features, which are then aligned with the dimensions of the large language model through a multilayer perceptron (MLP) block.

\subsubsection{MiniCPM-o}
The MiniCPM-o \cite{yao2024minicpm} is an open-source multimodal large language model (MLLM) developed by OpenBMB, capable of processing image, text, audio, and video inputs and generating high-quality text and speech outputs in an end-to-end manner. The model is based on SigLip-400M, Whisper-medium-300M, and Qwen2.5-7B-Instruct with a total of 8B parameters. 

\begin{table*}[t]
    \caption{Hyperparameter setting of MLLMs.}
    \centering
    \begin{tabular}{lcccc}
            \toprule[2pt]
              & Training Epochs& Learning Rate & LoRA Rank & LoRA $\alpha$ \\
            \midrule[1pt]
            HumanOmni & 1 & 2e-5 & 128 & 256   \\
            Qwen2.5Omni & 1 & 1e-4 & 8 & 32   \\
            VideoLLaMA2-AV & 1 & 2e-5 & 128 & 256  \\ 
            Ola & 1 & 2e-5 & 128 & 256  \\
            MiniCPM-o & 1 & 1e-4 & 8 & 32  \\
            Emotion-LLaMA & 10 to 20 & 1e-6 & 64 & 16 \\
            PandaGPT & 5 to 10 & 5e-4 & 32 & 32 \\
            \bottomrule[2pt]
    \end{tabular}
    \label{tab:hyperparameter}
\end{table*}

\begin{table}[t]
    \caption{Statistics of datasets in the benchmark.}
    \centering
    \begin{tabular}{lccccc}
            \toprule[2pt]
            Dataset & Type & \# Train & \# Valid & \# Test  \\
            \midrule[1pt]
            CMU-MOSI & MSA & 1281 & 229 & 685   \\
            CMU-MOSEI & MSA & 16326 & 1871 & 4659   \\
            CH-SIMS & MSA & 1368 & 456 & 457  \\ 
            CH-SIMS v2 & MSA & 2722 & 647 & 1034  \\
            MELD & MER & 9989 & 1109 & 2610  \\
            UR-FUNNY v2 & MHD & 7614 & 980 & 994  \\
            \bottomrule[2pt]
    \end{tabular}
    \label{tab:dataset_statistic}
\end{table}

\subsubsection{PandaGPT}
The PandaGPT \cite{su2023pandagpt} is a groundbreaking multimodal model capable of processing six modalities, including image/video, text, audio, depth, thermal, and inertial measurement units, while generating text responses. Its core architecture combines the multimodal encoders from ImageBind \cite{girdhar2023imagebind} and the LLM from Vicuna, creating a system for vision- and audio-grounded instruction following tasks.

\subsubsection{Emotion-LLaMA}
The Emotion-LLaMA \cite{cheng2024emotion} is a multimodal large language model designed for accurate emotion recognition and reasoning. The model integrates audio, visual, and textual inputs through emotion-specific encoders and employs instruction tuning on the MERR dataset \cite{cheng2024emotion} to enhance emotional recognition and reasoning capabilities.

The audio encoder employs HuBERT \cite{hsu2021hubert}, while the visual encoder uses a combination of MAE (Masked Autoencoders) \cite{sun2023mae}, VideoMAE (Masked Autoencoders for video) \cite{tong2022videomae}, and EVA (Efficient Vision Analysis) \cite{fang2023eva} to capture facial details, dynamics, and context. The multimodal features are aligned into a shared space using a modified LLaMA language model \cite{touvron2023llama}, which processes these inputs through a structured prompt template.

\subsection{Statistics of Datasets}
In this study, we employ six datasets, encompassing multimodal sentiment analysis (MSA) datasets (CMU-MOSI \cite{zadeh2016mosi}, CMU-MOSEI \cite{zadeh2018multimodal}, CH-SIMS \cite{yu2020ch}, and CH-SIMS v2 \cite{liu2022make}), multimodal emotion recognition (MER) dataset (MELD \cite{poria2018meld}), and multimodal humor detection (MHD) dataset (UR-FUNNY v2 \cite{hasan2019ur}). Here, We present a concise overview of these datasets below, with detailed statistics summarized in the table \ref{tab:dataset_statistic}.

\subsection{Hyperparameter Setting}
In terms of hyperparameter settings, the training epoch for Emotion-LLaMA was selected from {10,20,30,40}, while that for PandaGPT was chosen within the range of 1 to 10. For the remaining models, the training epoch was selected from {1,2,3}. The learning rate of the models was adjusted within the range of 1e-6 to 1e-3. For the LoRA module, the rank and $\alpha$ parameters were set to {8, 16, 64, 128, 256} and {16, 32, 128, 256, 512}, respectively. Please refer to Table \ref{tab:hyperparameter} for detailed information on the hyperparameter settings employed in our experiments.

\vfill

\end{document}